\documentclass[preprint,12pt]{elsarticle}

\usepackage{graphicx}
\usepackage{amssymb}
\usepackage{amsmath}
\usepackage{mathtools}
\DeclareMathOperator*{\argmax}{arg\,max}

\usepackage{hyperref}
\usepackage{lineno}
\usepackage{boldcommands}

\usepackage{lscape}

\journal{NeuroImage}

\begin{document}

\begin{frontmatter}

%% Title, authors and addresses

\title{Fast Infant MRI Skullstripping with Multiview 2D Convolutional Neural Networks}

%% use the tnoteref command within \title for footnotes;
%% use the tnotetext command for the associated footnote;
%% use the fnref command within \author or \address for footnotes;
%% use the fntext command for the associated footnote;
%% use the corref command within \author for corresponding author footnotes;
%% use the cortext command for the associated footnote;
%% use the ead command for the email address,
%% and the form \ead[url] for the home page:
%%
%% \title{Title\tnoteref{label1}}
%% \tnotetext[label1]{}
%% \author{Name\corref{cor1}\fnref{label2}}
%% \ead{email address}
%% \ead[url]{home page}
%% \fntext[label2]{}
%% \cortext[cor1]{}
%% \address{Address\fnref{label3}}
%% \fntext[label3]{}

%% use optional labels to link authors explicitly to addresses:
%% \author[label1,label2]{<author name>}
%% \address[label1]{<address>}
%% \address[label2]{<address>}

\author[mgh]{Amod Jog}
\author[bch]{P. Ellen Grant}
\author[wayne,uct,uct2]{Sandra W. Jacobson}
\author[uct,uct2]{Joseph L. Jacobson}
\author[mgh]{Andre van der Kouwe}
\author[uct]{Ernesta M. Meintjes}
\author[mgh,mit]{Bruce Fischl}
\author[mgh]{Lilla Z\"ollei}
\ead[mgh]{lzollei@nmr.mgh.harvard.edu}
\address[mgh]{Athinoula A. Martinos Center for Biomedical Imaging, Department of Radiology, Massachusetts General Hospital and Harvard Medical School}
\address[mit]{MIT Computer Science and Artificial Intelligence Laboratory}
\address[bch]{Fetal Neonatal Neuroimaging and Developmental Science Center, Division of Newborn Medicine, Boston Children’s Hospital and Harvard Medical School}
\address[uct]{Department of Human Biology, University of Cape Town Faculty of Health Sciences, Cape Town, South Africa}
\address[uct2]{Department of Psychiatry and Mental Health, University of Cape Town Faculty of Health Sciences, Cape Town, South Africa}
\address[wayne]{Department of Psychiatry and Behavioral Neurosciences, Wayne State University School of Medicine, Detroit, USA}

\begin{abstract}
%% Text of abstract
Skullstripping is defined as the task of segmenting brain tissue from a
full head magnetic resonance image~(MRI). It is a critical component in neuroimage processing pipelines. Downstream deformable registration and
whole brain segmentation performance is highly dependent on accurate
skullstripping. Skullstripping is an especially challenging task for
infant~(age range 0--18 months) head MRI images due to the significant size and shape variability of the head and the brain in that age range. Infant brain
tissue development also changes the $T_1$-weighted image contrast over time,
making consistent skullstripping a difficult task. Existing tools for adult brain MRI skullstripping are ill equipped to handle these variations and a
specialized infant MRI skullstripping algorithm is necessary. In this paper, we
describe a supervised skullstripping algorithm that utilizes three trained
fully convolutional neural networks~(CNN), each of which segments 2D $T_1$-weighted slices in axial, coronal, and sagittal views respectively. The three
probabilistic segmentations in the three views are linearly fused and
thresholded to produce a final brain mask. We compared our method to existing
adult and infant skullstripping algorithms and showed significant
improvement based on Dice overlap metric~(average Dice of 0.97) with a manually
labeled ground truth data set. Label fusion experiments on multiple, unlabeled
data sets show that our method is consistent and has fewer failure modes. In
addition, our method is computationally very fast with a run time of 30
seconds per image on NVidia P40/P100/Quadro 4000 GPUs.

\end{abstract}

\begin{keyword}
infant \sep brain \sep MRI \sep skullstripping \sep deep learning
%% keywords here, in the form: keyword \sep keyword

%% MSC codes here, in the form: \MSC code \sep code
%% or \MSC[2008] code \sep code (2000 is the default)

\end{keyword}

\end{frontmatter}

%%
%% Start line numbering here if you want
%%

%% main text
\section{Introduction}
\label{sec:intro}
Skullstripping is a critical image preprocessing step in most neuroimage
processing pipelines~\cite{fischl2004aseg}.
The goal of skullstripping is to take an input whole head magnetic resonance
image~(MRI) and output a binary mask with a value of one for the brain and
zero for all the extra-cerebral tissues such as skin, muscle, fat, bone etc.
Extra-cerebral tissues significantly vary in their size and shape across subjects. Pulse sequences for brain imaging are also typically optimized to provide the best image contrast for brain tissues such as gray~(GM) and white matter~(WM). This results in extra-cerebral tissues with large
variation in their intensities and imaging features that can prove difficult to model in downstream whole brain segmentation and deformable registration
algorithms~\cite{ou2014tmi}, potentially reducing their accuracy.
Thus, skullstripping becomes a necessary step for most neuroimage processing
pipelines to mask out uninteresting background and help increase the accuracy
of the complete segmentation and cortical reconstruction pipeline~\citep{fischl2004aseg}.

\begin{figure}[!htbp] \tabcolsep 1pt
	\centerline{
		\begin{tabular}{ccc}
      (a) Newborn &  (b) 6 months & (c) 18 months \\
			\includegraphics[width=.13\textwidth,height=.15\textwidth]{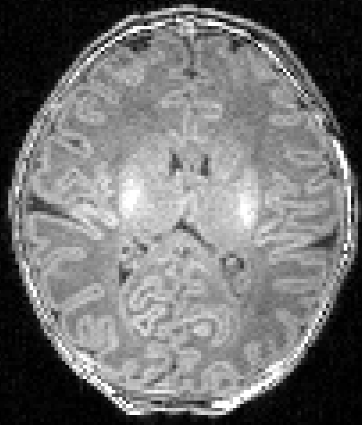} &
			\includegraphics[width=.13\textwidth,height=.15\textwidth]{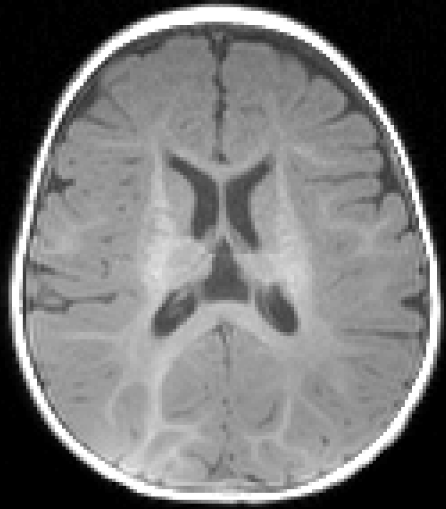} &
			\includegraphics[width=.13\textwidth,height=.15\textwidth]{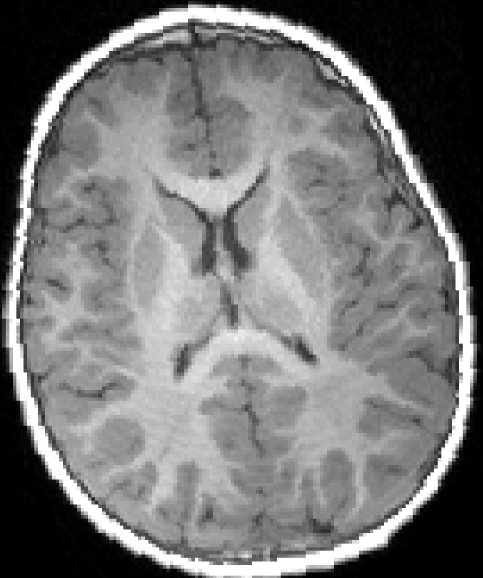}\\
      \includegraphics[width=.13\textwidth,height=.15\textwidth]{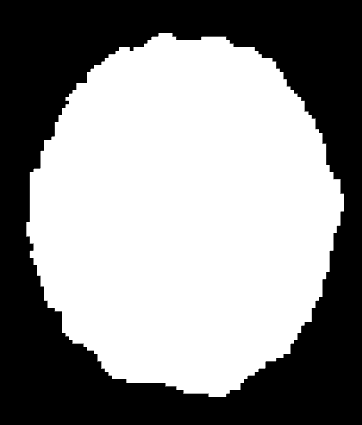} &
      \includegraphics[width=.13\textwidth,height=.15\textwidth]{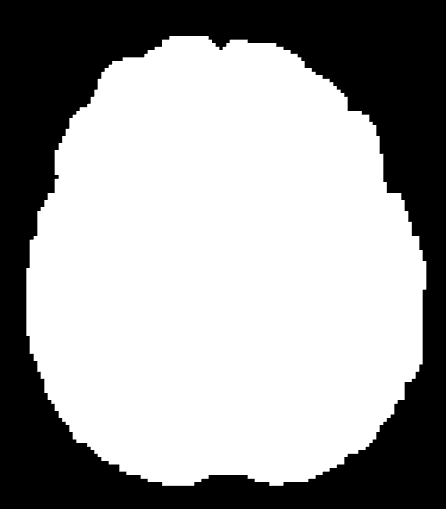} &
      \includegraphics[width=.13\textwidth,height=.15\textwidth]{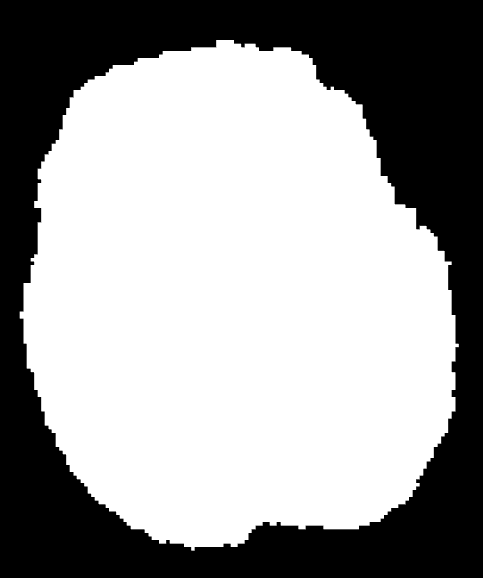} \\
      \includegraphics[width=.13\textwidth,height=.15\textwidth]{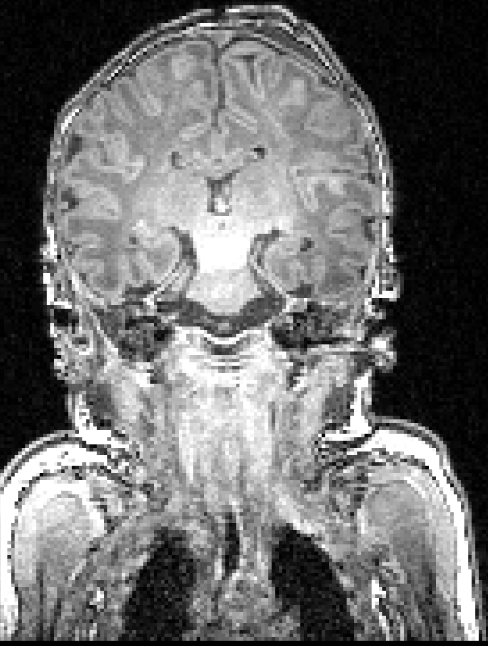} &
      \includegraphics[width=.13\textwidth,height=.15\textwidth]{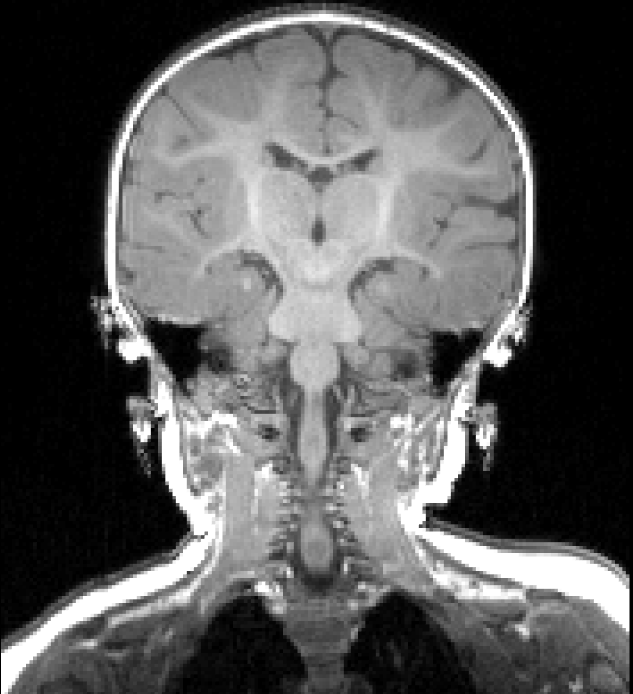} &
      \includegraphics[width=.13\textwidth,height=.15\textwidth]{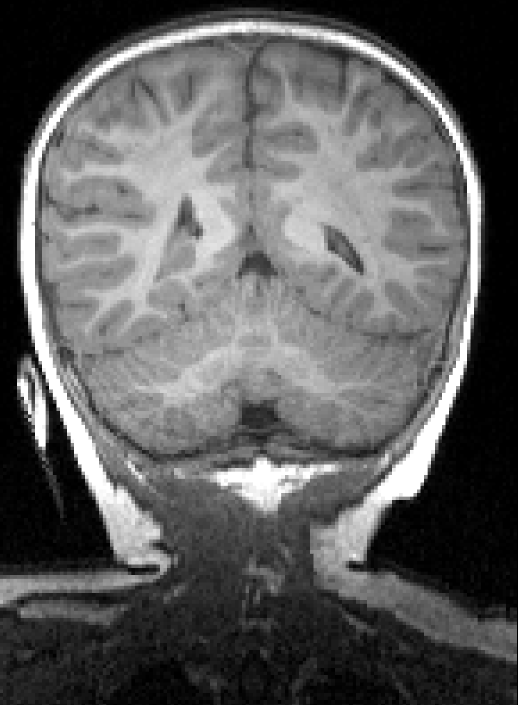} \\
      \includegraphics[width=.13\textwidth,height=.15\textwidth]{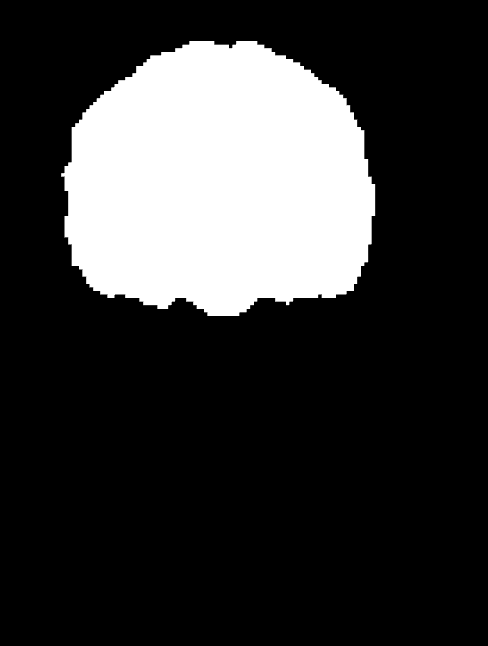} &
      \includegraphics[width=.13\textwidth,height=.15\textwidth]{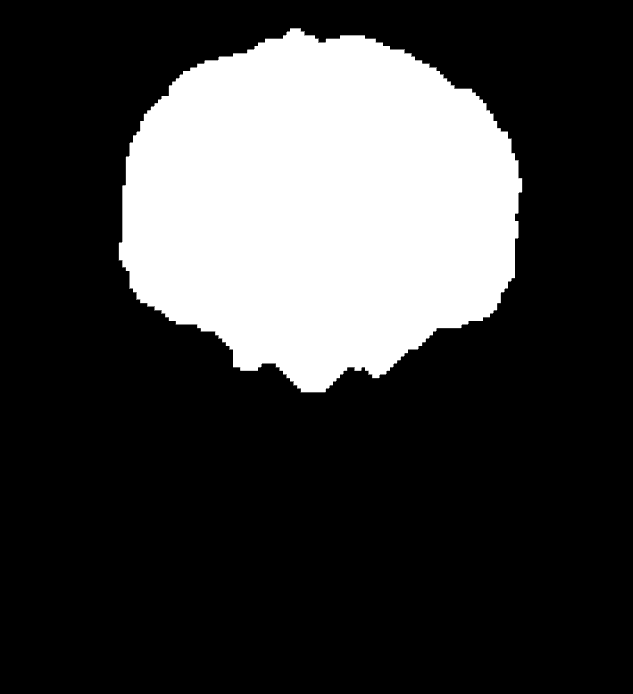} &
      \includegraphics[width=.13\textwidth,height=.15\textwidth]{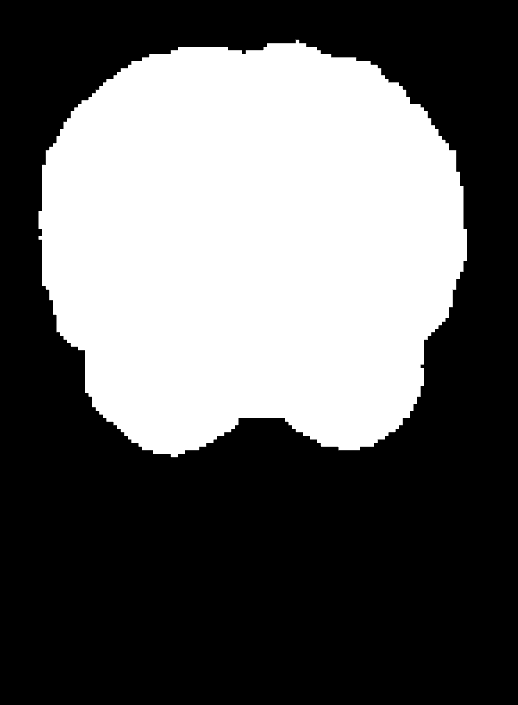} \\
      \includegraphics[width=.13\textwidth,height=.15\textwidth]{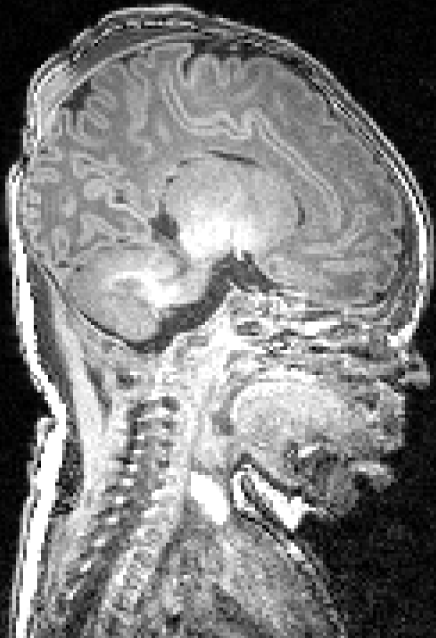} &
      \includegraphics[width=.13\textwidth,height=.15\textwidth]{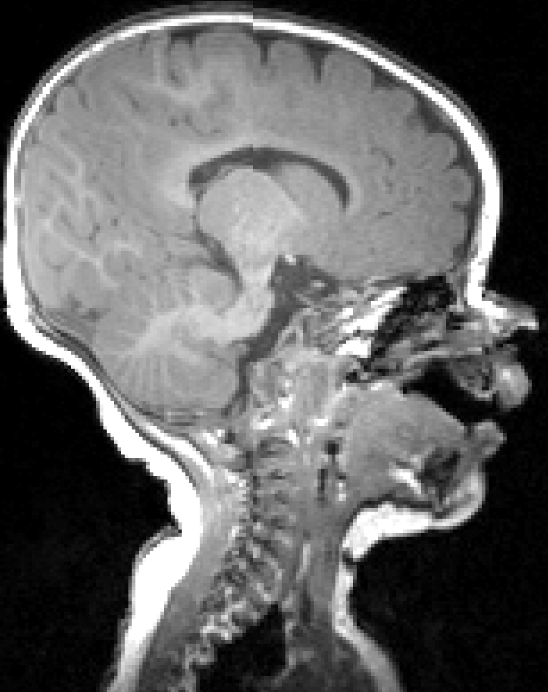} &
      \includegraphics[width=.13\textwidth,height=.15\textwidth]{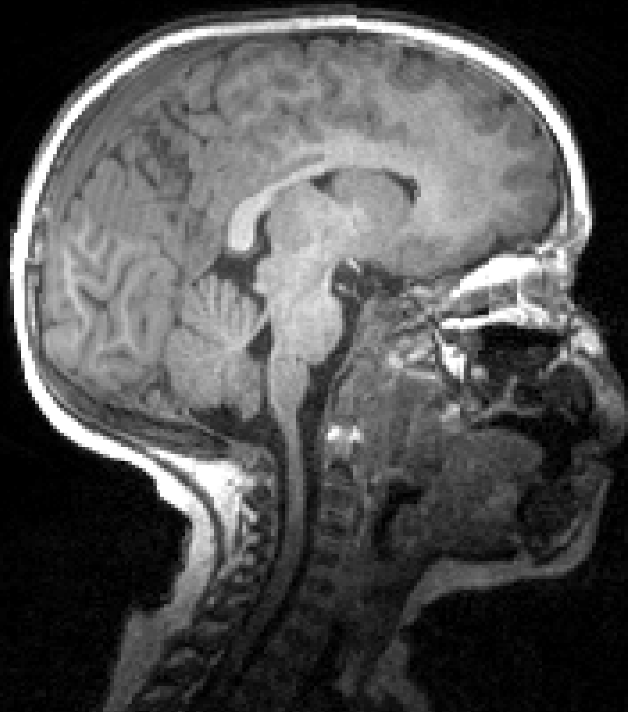} \\
      \includegraphics[width=.13\textwidth,height=.15\textwidth]{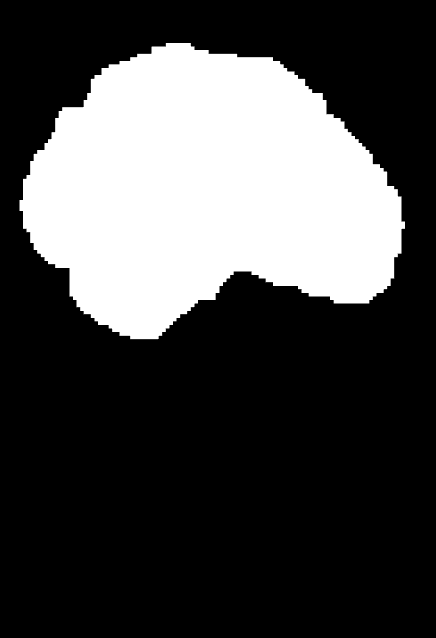} &
      \includegraphics[width=.13\textwidth,height=.15\textwidth]{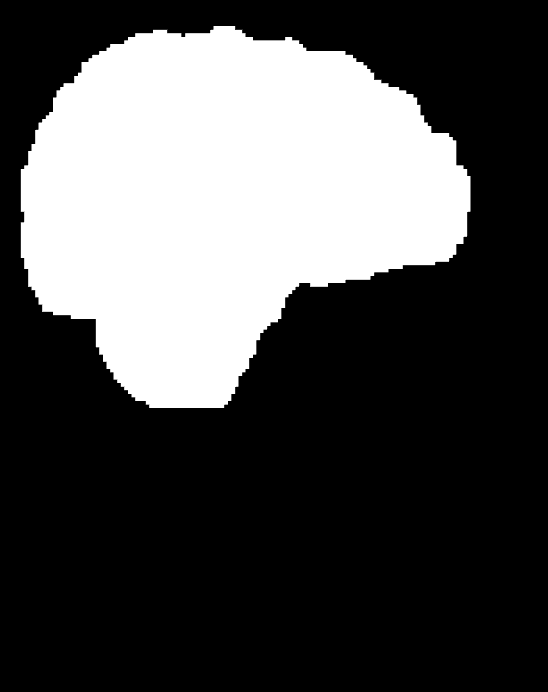} &
      \includegraphics[width=.13\textwidth,height=.15\textwidth]{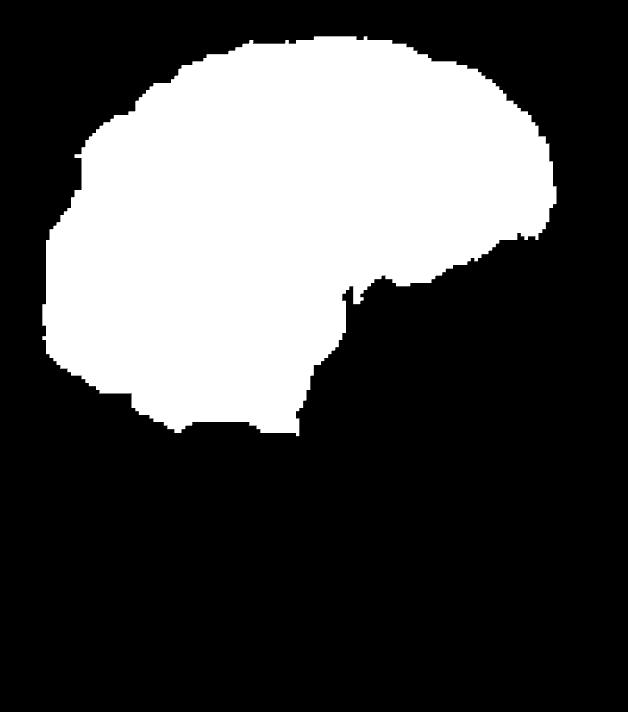} \\
	\end{tabular}
	}
	\caption{Full head MPRAGE acquisitions and manually labeled skull masks for axial (rows 1 and 2), coronal (row 3 and 4), and sagittal (rows 5 and 6) of infants of (a) newborn, (b) 6 months and (c) 18 months of age.}
	\label{fig:data}
\end{figure}

Skullstripping is all the more important when processing infant brain MRI.
Infant heads almost double in size in the first two years of life, as opposed to adult brains, which are of a relatively fixed size. Moreover, tissue contrast in the infant brain changes significantly
with different stages of development. In Fig.~\ref{fig:data} we show MPRAGE (magnetization prepared gradient echo) $T_1$-weighted acquisitions \cite{Mugler:1991} of three different infants in three stages of development: newborn (column 1), 6 months old~(column 2), and 18 months
old~(column 3). The tissue contrast differs significantly in all these stages. The small dimensions of the infant head can also result in variations in the field of view and scanner distortions, as is apparent in Fig.~\ref{fig:data}.

There has been a significant amount of work in the development of
skullstripping algorithms for adult head MRI. These include
methods using spherical expansion~\cite{cox1996afni}, filtering and edge-detection-based
approaches~\cite{shattuck2001bes}, deformable model-based segmentation~\cite{smith2002BET}, variations of watershed algorithms~\cite{hahn2000watershed,segonne2004hybrid,carass2011spectre}, a generative-discriminative framework~\cite{iglesias2011robex}, patch-based sparse
reconstruction methods for multi-modal data~\cite{roy2017monstr}, multi-atlas registration-based skullstripping~\cite{doshi2013mass} among many others.
These methods tend to have a sub-optimal performance for infant MRI skullstripping
due to inherent assumptions about skull shape, size, and appearance in adults~\cite{mahapatra2012graph}.
Infant MRI skullstripping has received some attention in recent years.
Methods have been developed to work simultaneously for pediatric and adult brain MRI~\cite{zhuang2006ni,chiverton2007cbm,eskildsen2012ni}.
These include model-based level set driven skullstripping~\cite{zhuang2006ni}, a statistical morphology-based tool~\cite{chiverton2007cbm}, and non-local patch-based segmentation~\cite{eskildsen2012ni}.
A specialized method for infant skullstripping using the prior shape of neonatal brains that is learned using a labeled set of atlas images, with final
segmentation using graph cuts was described by~\cite{mahapatra2012graph}. A meta-learning algorithm that combines
the outputs of brain extraction tool~(BET)~\cite{smith2002BET} and brain
surface extractor~(BSE) was described in~\cite{shattuck2001bes}. A multi-atlas registration-based
method focusing on apparent diffusion coefficient~(ADC) images of infants has also been developed~\cite{Ou2015}.
Recently there have been methods that post-process the BET outputs
for infant data specifically~\cite{alansary2016jbhi}.
Multi-atlas registration-based methods tend to be
computationally expensive due to the many registrations needed to align the atlases to the subject
image. The same holds true for meta learning-based methods that need to run
multiple available skullstripping algorithms and combine their results.
Intensity and morphology-based methods can also take up to 10--20 minutes to produce a brain mask. Segmentation accuracy can also reduce in the presence of intensity inhomogeneities that can vary the intensity gradient magnitudes between similar tissues across slices.

Recently, deep learning architectures based on convolutional neural networks~(CNN) have been successfully applied to a variety of
medical image segmentation problems~\cite{roth2015deeporgan, kamnitsas2017, niftynet18}.
A 3D deep learning framework for skullstripping was described by~\cite{kleesiek2016deep} that used 3D patches
of size $65\times65\times65$ as input to the network and the output was $7\times7\times7$-sized patches with
probability values of each voxel belonging to the brain class. A great advantage of deep learning algorithms is
the fast inference time. Inference using CNNs is very fast and fast skullstripping can significantly speed up neuroimage
processing pipelines that typically take hours to complete for a single subject volume.

In this work, we describe SkullStripping CNN or \textit{SSCNN}, a 2D multi-view
CNN-based skullstripping approach. We train three independent
2D CNNs, one each for 2D slices in coronal, axial, and sagittal views. Predictions of these three networks
are linearly combined to produce a final 3D mask. Past works have shown that
using multi-view approaches can reduce disambiguities as compared to a single view 2D segmentation~\cite{bekker2016multiview}.
2D multi-view CNNs have also been used for whole brain segmentation~\cite{guharoy2019quicknat}. 2D CNNs are more preferable
for the skullstripping task because even for a human labeler, skullstripping is a visually intuitive task to perform on 2D slices instead of 3D patches.
Errors made in a single view prediction can potentially be corrected by predictions from the other views. Additionally, skullstripping is usually one of the first preprocessing tasks for brain MRI and the images can present with uncorrected intensity inhomogeneities. The intensity bias is usually assumed
to be a slow-changing multiplicative field. A multiview 2D CNN segments each
slice independently, for each of the three independent views. Intensity
inhomogeneity may adversely affect segmentation of slices in one of the views but the other views can compensate for that loss. In contrast, a 3D CNN would segment a 3D patch, which can have varying intensity inside it making the segmentation less robust.
Segmenting a full 3D image instead of 3D patches is ideal but is not possible
to achieve using CNNs due to GPU memory constraints.

Our paper is organized as follows. Section~\ref{sec:method} describes the method, training, and prediction
using multi-view 2D CNNs. In Section~\ref{sec:experiments}, we describe parameter selection experiments
and comparison with other skullstripping algorithms. Finally, in Section~\ref{sec:conclusion}, we summarize
our observations and conclude with possible avenues of future development.

\section{Method}
\label{sec:method}

\begin{figure}[!htbp]
\begin{tabular}{cc}
	\includegraphics[width=.45\textwidth]{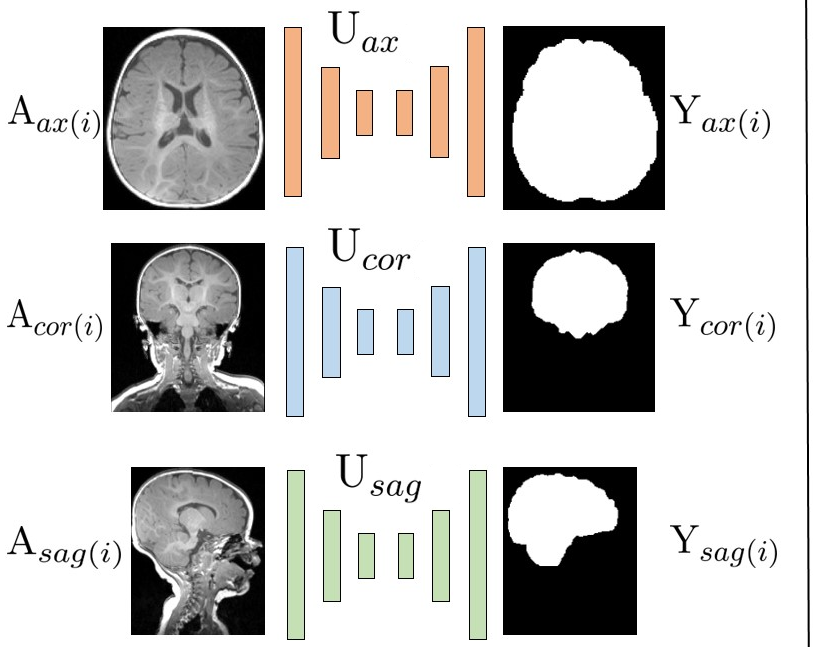} &
	\includegraphics[width=.45\textwidth]{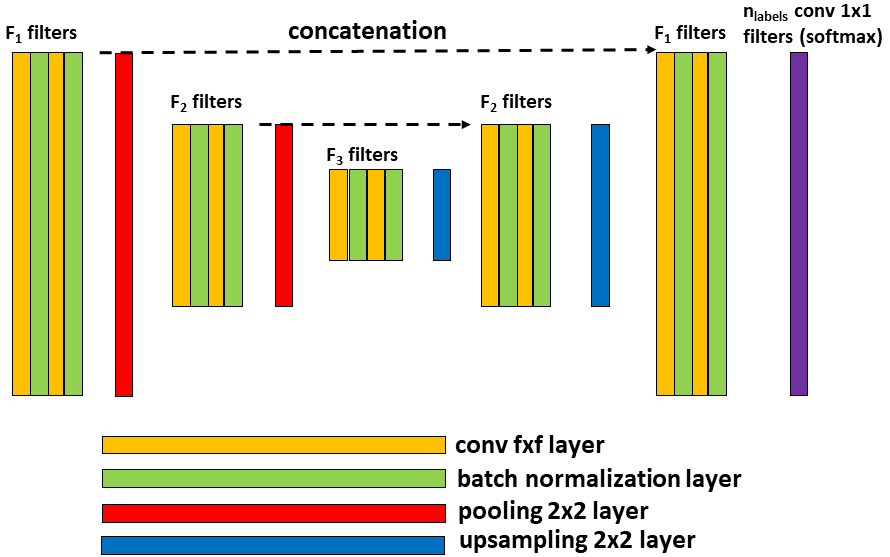}
\end{tabular}
\caption{The left half of the figure shows training of 3 2D
U-Net segmentation architectures, each for slices oriented in
one of the three cardinal orientations coronal, sagittal,
and axial. The U-net architecture for all three with $D_l=3$ is the one
as shown on the right.}
\label{fig:method}
\end{figure}

Let $\mathcal{A} = \{A_{(1)}, A_{(2)}, \ldots, A_{(M)}\}$ be a collection of
$M$ images with a paired expert manually labeled image set
$\mathcal{Y} = \{Y_{(1)}, Y_{(2)}, \ldots Y_{(M)}\}$. Each $A_{(i)} \in \mathcal{A}$ is a 3D image of size $256\times256\times256$ and is resampled to
a voxel size of $1\times1\times1$~mm$^3$. The paired collection
$\{\mathcal{A}, \mathcal{Y}\}$ is referred to as the training image set.
Figures~\ref{fig:data}(a)--(f) show three training data subjects in three age
ranges and their manually labeled brain masks.
From $\{\mathcal{A}, \mathcal{Y}\}$, we extract corresponding 2D slices
in the coronal orientation of size $256\times256$ to create a 2D
training data set $\{\mathcal{A}_{cor}, \mathcal{Y}_{cor}\}$. Similarly,
we extract slices from the sagittal and axial orientations from $\{\mathcal{A}, \mathcal{Y}\}$ to generate 2D training data sets, $\{\mathcal{A}_{sag}, \mathcal{Y}_{sag}\}$, and $\{\mathcal{A}_{ax}, \mathcal{Y}_{ax}\}$.
Figure~\ref{fig:method}~(left) shows training examples for all three orientations.

\subsection{SSCNN: Network Architecture and Training}
\label{sec:training}
For each orientation $r \in \{ax, cor, sag \}$, we train a 2D fully convolutional CNN $U_{r}$ that takes as
input the $T_1$-weighted 2D slice in that orientation to predict the corresponding manually labeled slice. Paired intensity and label slice data $\{ \mathcal{A}_{r}, \mathcal{Y}_{r} \}$ is used as training data
for the CNN $U_{r}$. We are interested in the semantic voxel-wise segmentation of the input 2D slice
and therefore use the well-known U-Net architecture~\cite{Ronneberger:2015} for $U_{r}$.

The U-Net consists of an encoder block with $L$ levels, followed by a symmetric
decoder block. Figure~\ref{fig:method}~(right) shows an example U-net architecture
with $L=3$ levels. In each level $l$ of the encoder block, we have $D_l$ convolutional
layers~(orange layer in Fig.~\ref{fig:method}). The number of filters in each
convolutional layer at level $l$ is $F_l$ and the number of filters in subsequent levels is $F_l  = 2F_{l-1}$.  All convolutional layers, except the last layer, have a rectified linear~(ReLU)
activation, with a filter size $f\times f$. We choose the optimal values for $L$, $f$, $F_1$, and $D_1$
by cross-validation experiments. Each convolutional layer~(conv) is followed by a
batch normalization~(BN) layer~\cite{Ioffe:2015}~(green in Fig.~\ref{fig:method}).
At the end of each level (except the last one) is a pooling layer~(red in Fig.~\ref{fig:method})
that downsamples the output of that level by 2 in the image height-width directions.
At the deepest level, the encoder block ends and the decoder block begins.
In the decoder, the deepest conv+BN layers are followed by an upsampling layer~(blue in Fig.~\ref{fig:method})
that upsamples the output of the previous level by a factor of 2 in the image height-width directions.
The upsampled output is followed by a symmetric number of levels as the encoder.
The number of filters and number of conv+BN layers at each level of the decoder block
is the same as the corresponding symmetric level on the encoder block.
The final layer consists of $n_{labels} = 2$, $1\times1$ conv layers with a softmax activation.
To summarize, when a network $U_{r}$ is applied to an input slice of the shape $(256,256,1)$ in orientation $r$, the output is of the shape $(256,256,2)$, and stores the probabilities of each voxel in the input slice belonging to the background (class 0), or brain~(class 1).

All three U-Nets are trained to minimize a soft Dice-based loss averaged over the
whole batch. The loss is shown in Eqn.~\ref{eq:loss}, where $v$ denotes the voxels
present in ground truth slice $\boldy_{true}(b)$ and predicted slice $\boldy_{pred}(b)$ of the
$b^\textrm{th}$ training sample in a batch of $N_B$ samples in $\boldy_{true}$
and $\boldy_{pred}$. During training, $\boldy_{pred}(b)$ is a $256\times256\times n_{labels}$-sized
tensor with each voxel storing the softmax probability of it belonging to a particular label.
$\boldy_{true}(b)$ is a similarly-shaped one-hot encoding of the label present at each of the voxels.
\begin{equation}
\textrm{Loss}(\boldy_{true}, \boldy_{pred}) = \frac{1}{N_B}\sum_{b}\left(1 - \frac{2\sum _{v}\boldy_{true}(b)\cdot\boldy_{pred}(b)}{\sum_{v}(||\boldy_{true}(b)||^2 + ||\boldy_{pred}(b)||^2 )}\right)
\label{eq:loss}
\end{equation}
We use the Adam optimization algorithm~\cite{Kingma:2014} to minimize the loss with an initial
learning rate of $10^{-5}$ that reduces by half if the training loss does
not reduce for five continuous epochs. Each epoch
uses 3000 training slice pairs extracted from 23 training subjects.
We use $\approx$600 slices extracted from a validation dataset that
consists of three subjects. We train each of the $U_{cor}$, $U_{sag}$, and $U_{ax}$
for 30 epochs. The training loss reduces monotonously as the number of epochs
increases, however, we select the model with the minimum validation loss as our
final trained model. This avoids using networks that overfit to the
training data and do not generalize well for test data from other sources.
All three networks $U_{cor}$, $U_{sag}$, $U_{ax}$ are trained independent of each other.

\subsection{SSCNN: Prediction}
\label{sec:prediction}
\begin{figure}[!htbp]
\begin{tabular}{c}
	\includegraphics[width=.95\textwidth]{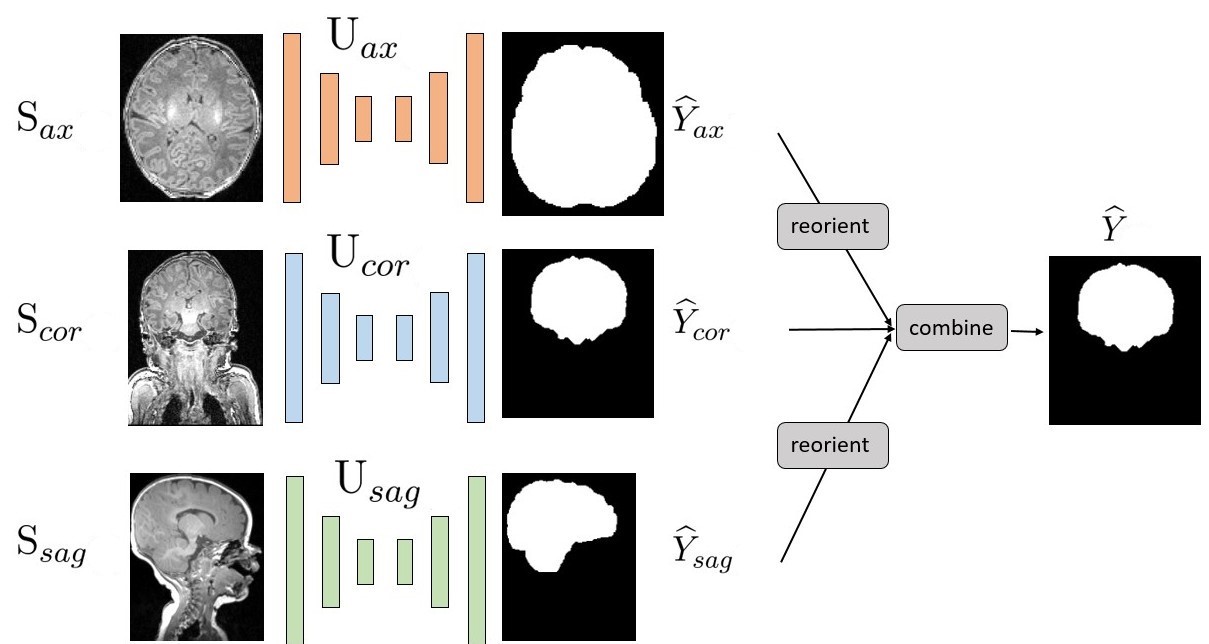}
\end{tabular}
\caption{The prediction workflow for SSCNN.}
\label{fig:prediction}
\end{figure}
Given a 3D test subject volume $S$, we reorient it in coronal, sagittal,
and axial directions to create 3D volumes $S_{cor}$, $S_{sag}$,
and $S_{ax}$ respectively. Let $S_{r}(z)$ be the $z^{\textrm{th}}$ slice
in $S_r$, $r \in \{ax, cor, sag\}$. We apply the trained network $U_r$ to
$S_{r}(z)$ to generate output $\hat{Y}_{r}(z) = U_{r}(S_{r}(z))$ for all $z$
to form $\hat{Y}_{r}$

Each of $\hat{Y}_{ax}$, $\hat{Y}_{cor}$, $\hat{Y}_{sag}$
is reoriented to the coronal orientation to create $\hat{Y}_{cor-ax}$, $\hat{Y}_{cor-cor}$, $\hat{Y}_{cor-sag}$,
respectively. Note that each $\hat{Y}_{cor-r}$ image stores the voxel-wise class probabilities.
We linearly combine these probability images to generate a fused result in the
coronal orientation, as shown in Eqn.~\ref{eq:lincombine}:
\begin{equation}
    \hat{Y} = w_{ax}\hat{Y}_{cor-ax} + w_{cor}\hat{Y}_{cor-cor} + w_{sag}\hat{Y}_{cor-sag}.
    \label{eq:lincombine}
\end{equation}
The weights are constrained such that $w_{ax} + w_{cor} + w_{sag} = 1$. We found the optimal weights
using cross-validation experiments to be $w_{ax} =  0.44$, $w_{cor} = 0.33$, and $w_{sag} = 0.23$.
Figure~\ref{fig:prediction} illustrates the SSCNN prediction workflow.
The brainmask is obtained by identifying the class with the higher probability value by
$\hat{Y}_{hard} =  \argmax_{\{0,1\}} \hat{Y}$. To remove any small non-brain blobs,
we identify the connected components in the binary labeled image and select the largest connected component
as the final labeled brain.

\section{Experiments}
\label{sec:experiments}
In this section we describe the training data set~(Section~\ref{sec:trainingdata}), the SSCNN parameter selection step~(Section~\ref{sec:parameters}), and a set of quantitative experiments demonstrating the high quality performance of our tool: a set of leave-one-out cross validation experiments and comparison with other skullstripping algorithms on the manually labeled training data set~(Section~\ref{sec:accuracy}), validation on an independent
manually labeled test data set~(Section~\ref{sec:uct_accuracy}), and a comparison on multiple unlabeled
data sets~(Section~\ref{sec:unlabeled_accuracy}).

\subsection{Training Data Set:}
\label{sec:trainingdata}
Our training data set was generated by using the volumes described in \cite{Macedo:2015}. A total of 26 T1-weighted MRI volumes, from subjects whose age is almost uniformly distributed in the age range of 0-18 months, together with their binary brain masks were used.

\subsection{SSCNN Parameter Selection}
\label{sec:parameters}
The 2D U-Nets used in SSCNN have a number of free parameters that need to be carefully selected for optimal performance. For parameter selection, we chose 13 subjects distributed roughly equally in the age range of 0--18 months as our parameter selection training set. We used four subjects to generate a validation data set to prevent overfitting. We tested the
trained network on the remaining nine subjects. While selecting the optimal value of a parameter, we iterated over a set of possible values by keeping the other parameters fixed and chose the parameter with the highest test segmentation accuracy over the nine subjects. The optimal parameters
are therefore dependent on the order in which they were fixed.

The parameters of interest for our proposed pipeline are the following: (1) kernel size~($f$), (2) depth of network~(number of pooling layers = $L$), (3) number
of filters in the first level~($F_1)$, and (4) number of convolutional layers in each level~($D_l$).

\subsubsection{Kernel Size~$(f)$}
\begin{figure}[!htbp]
    \centering
    \includegraphics[width=0.7\textwidth]{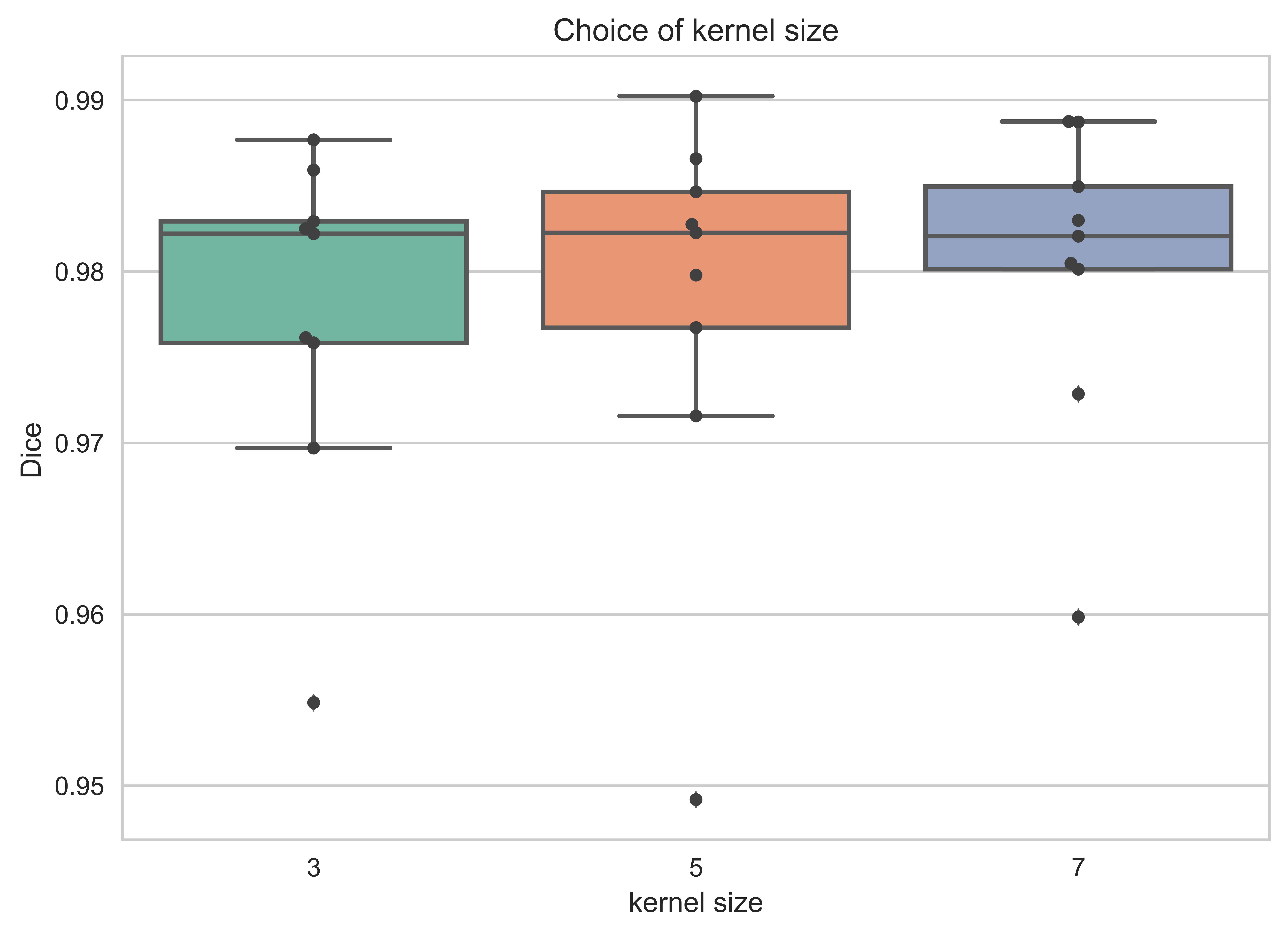}
    \caption{Dice coefficient with varying kernel size}
    \label{fig:kernel}
\end{figure}

This experiment was to select the most optimal kernel size of the 2D kernels used in each of the $U_r$ CNNs. Larger kernel sizes have larger receptive fields that can act on larger image contexts. Therefore we expect larger kernel sizes to improve the segmentation accuracy, but they also have a much
larger number of paramaters to learn. We experimented with $f = \{3, 5, 7\}$, by keeping the number of pooling layers $L = 6$, the number of filters in the first layer $F_1 = 16$, and the number of convolutional layers in each level $D_l=2$. Increasing the kernel size beyond $7\times7$ resulted in out of memory errors. Figure~\ref{fig:kernel} shows a boxplot of Dice overlap coefficients obtained on the nine test subjects when compared with the ground truth segmentation. The median Dice score increases as the kernel size increases,
with $f=7$ resulting in the highest value. Therefore, we fixed the filter shape to be $7\times7$.

\subsubsection{Depth of Network}
\begin{figure}[!htbp]
    \centering
    \includegraphics[width=0.7\textwidth]{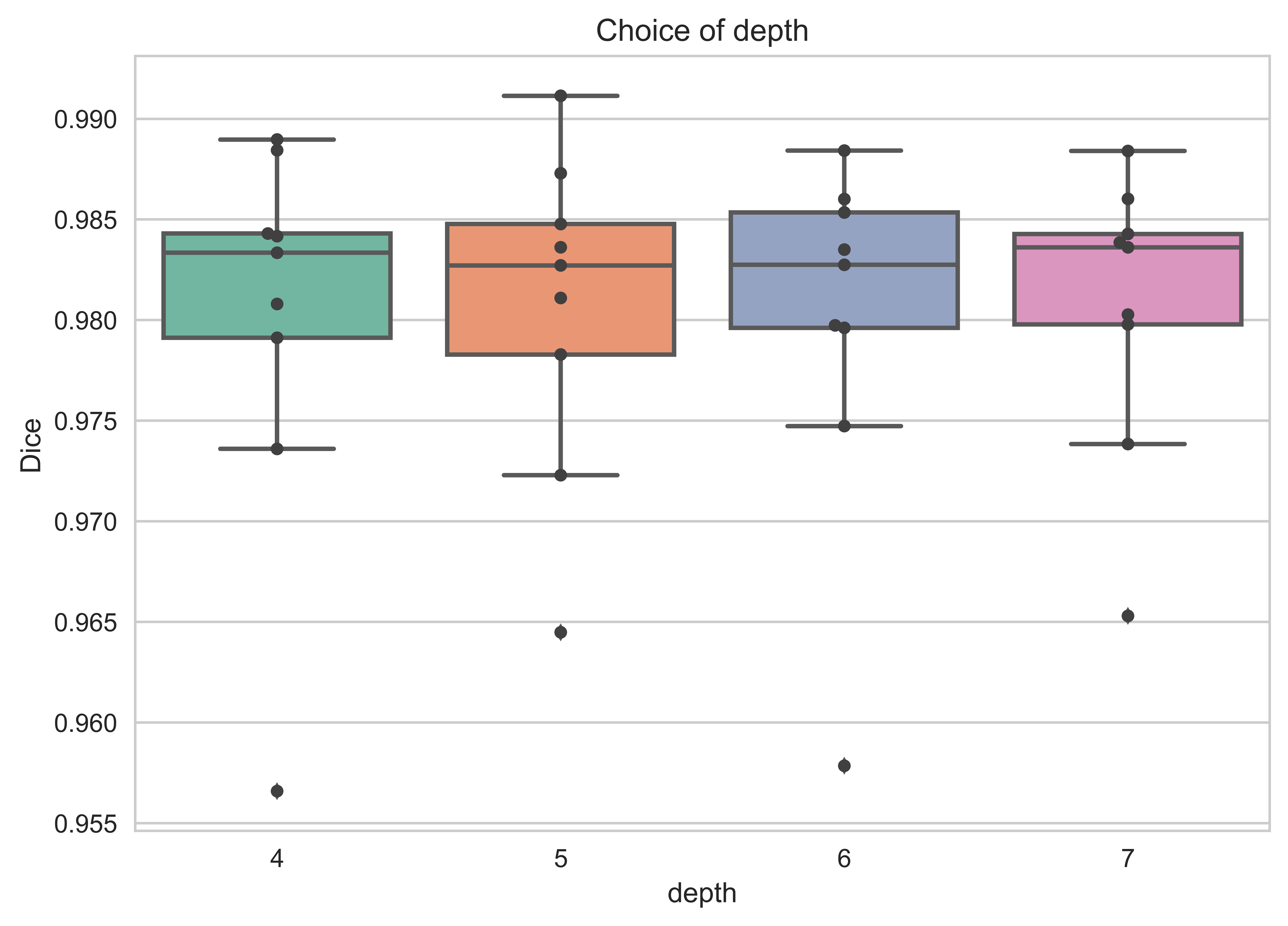}
    \caption{Dice coefficient with varying number of pooling layers}
    \label{fig:depth}
\end{figure}

In this experiment we varied the depth of the network given by $2L+1$,
where $L$ is the number of pooling layers in the network.
It has been observed that deeper networks lead to better performance.
We varied $L \in \{4, 5, 6, 7\}$ and kept $D_l=2$, $f=7$, and $F_1=16$.
In Fig.~\ref{fig:depth} we show boxplots for the four experiments,
where we observe that the depth of the network does not change segmentation accuracy as it increases from four to seven. We set $L=6$ as it produces marginally higher numbers than the rest, but it is not a significant difference.

\subsubsection{Number of Filters}
\begin{figure}[!htbp]
    \centering
    \includegraphics[width=0.7\textwidth]{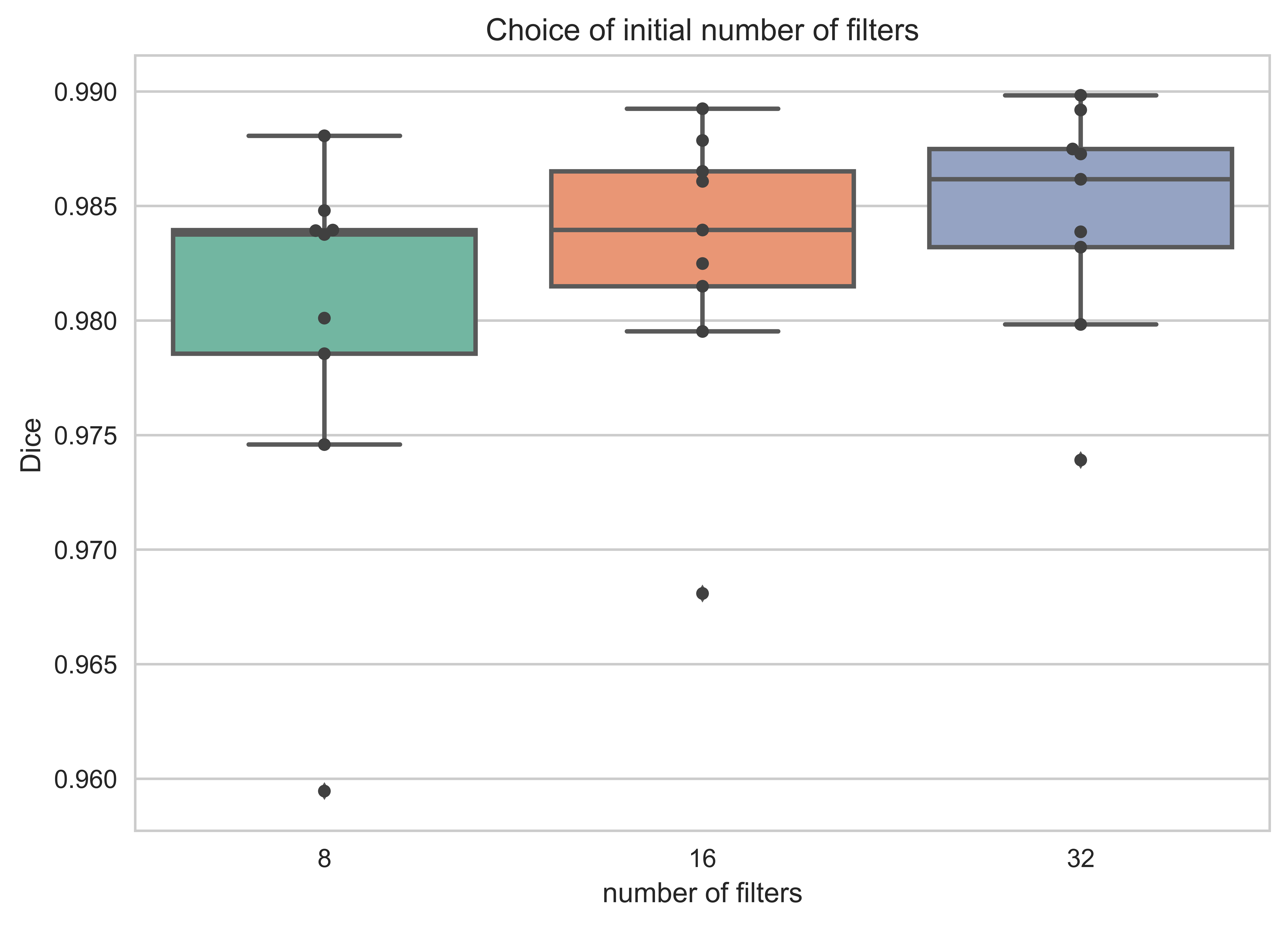}
    \caption{Dice coefficient with varying number of filters in the first level}
    \label{fig:numfilters}
\end{figure}

In this experiment, we chose the number of filters in the convolutional layers of the first
level $F_1$. As described in Section~\ref{sec:method}, the number of filters in the
subsequent levels $l$ is given by $F_l = 2F_{l-1}$, therefore the total number of
filters in the entire network is a function of the number of filters in the first level.
The more filters we have the more image variation can be captured at different
network depths. We expect that more filters in the network will lead to higher
segmentation accuracy.
We experimented with $F_1 \in \{8, 16, 32\}$, by keeping $L=6$, $f=7$, and $D_l = 2$. $F_1 > 32$ was not possible due to GPU memory constraints. Figure~\ref{fig:numfilters} shows a boxplot of Dice coefficients obtained on the nine
test subjects when compared with the ground truth segmentation.
The median Dice increases as $F_1$ increases from 8 to 32. $F_1=32$ produces the highest median Dice score and thus we set $F_1 = 32$.

\subsubsection{Depth per Level of Network}
\begin{figure}[!htbp]
    \centering
    \includegraphics[width=0.7\textwidth]{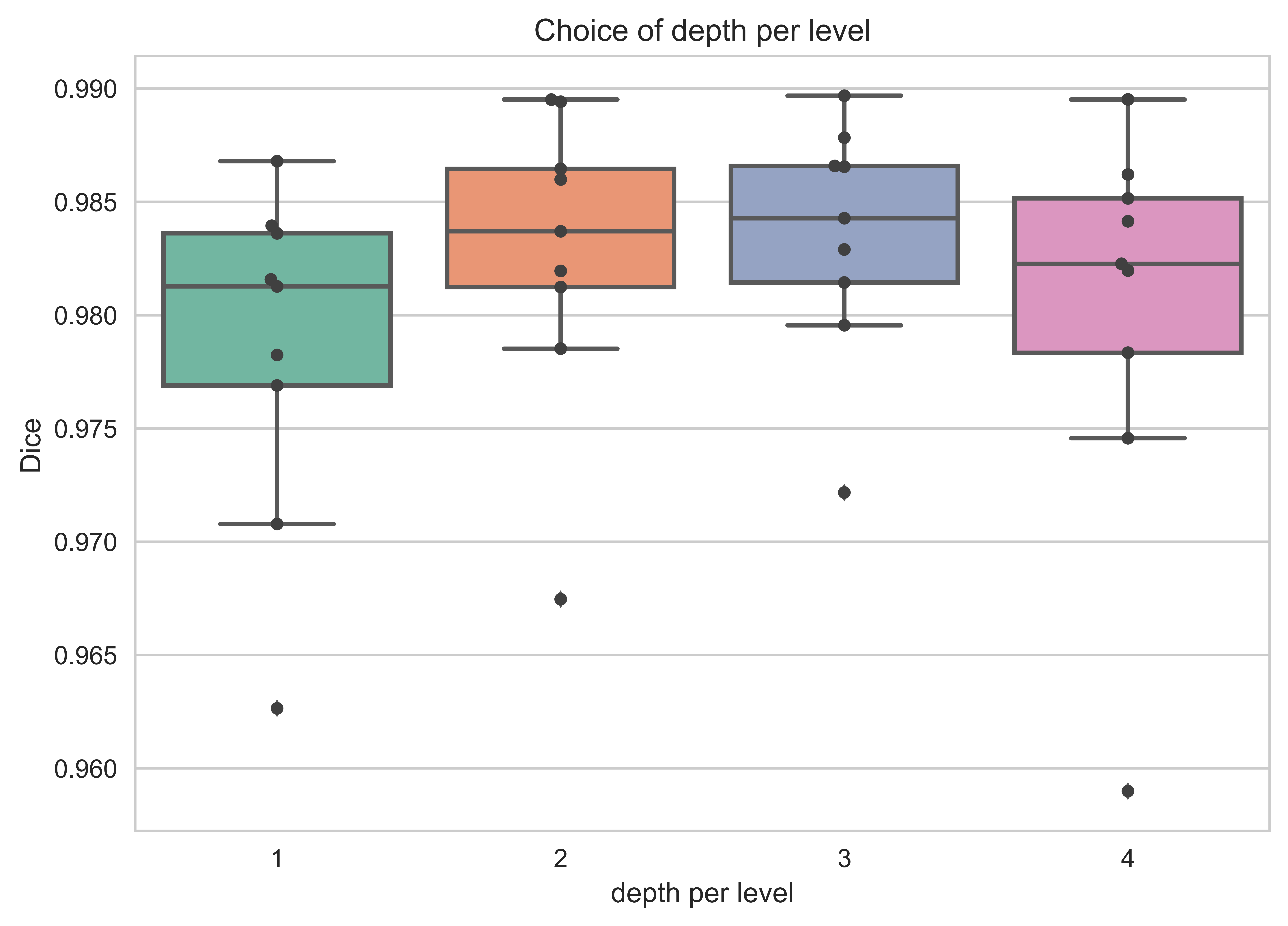}
    \caption{Dice coefficient with varying number of convolutional layers per level}
    \label{fig:depthperlevel}
\end{figure}

In these experiments, we varied the number of convolutional layers $D_l$ at each
level of the U-Net. We can expect that the higher the number of convolutional layers at each level, the
better the image texture is captured. However, this also increases the number of parameters
in their network leading to sub-optimal training and possible overfitting.
We varied $D_l \in \{1, 2, 3, 4\}$ by keeping $f=7$, $F_1=32$, $L=6$.
Figure~\ref{fig:depthperlevel} shows the Dice score boxplots calculated over the
nine test subjects. We observe that the median Dice score increases from $D_l=1$
to $D_l=3$ and then falls for $D_l=4$. Based on these experiments, we chose
$D_l=3$ for our optimal network.

\subsection{Evaluation of Segmentation Accuracy}
\label{sec:accuracy}
\subsubsection{Cross-validation on Training Data Set}
\begin{figure}[!htbp]
    \centering
    \includegraphics[width=0.8\textwidth]{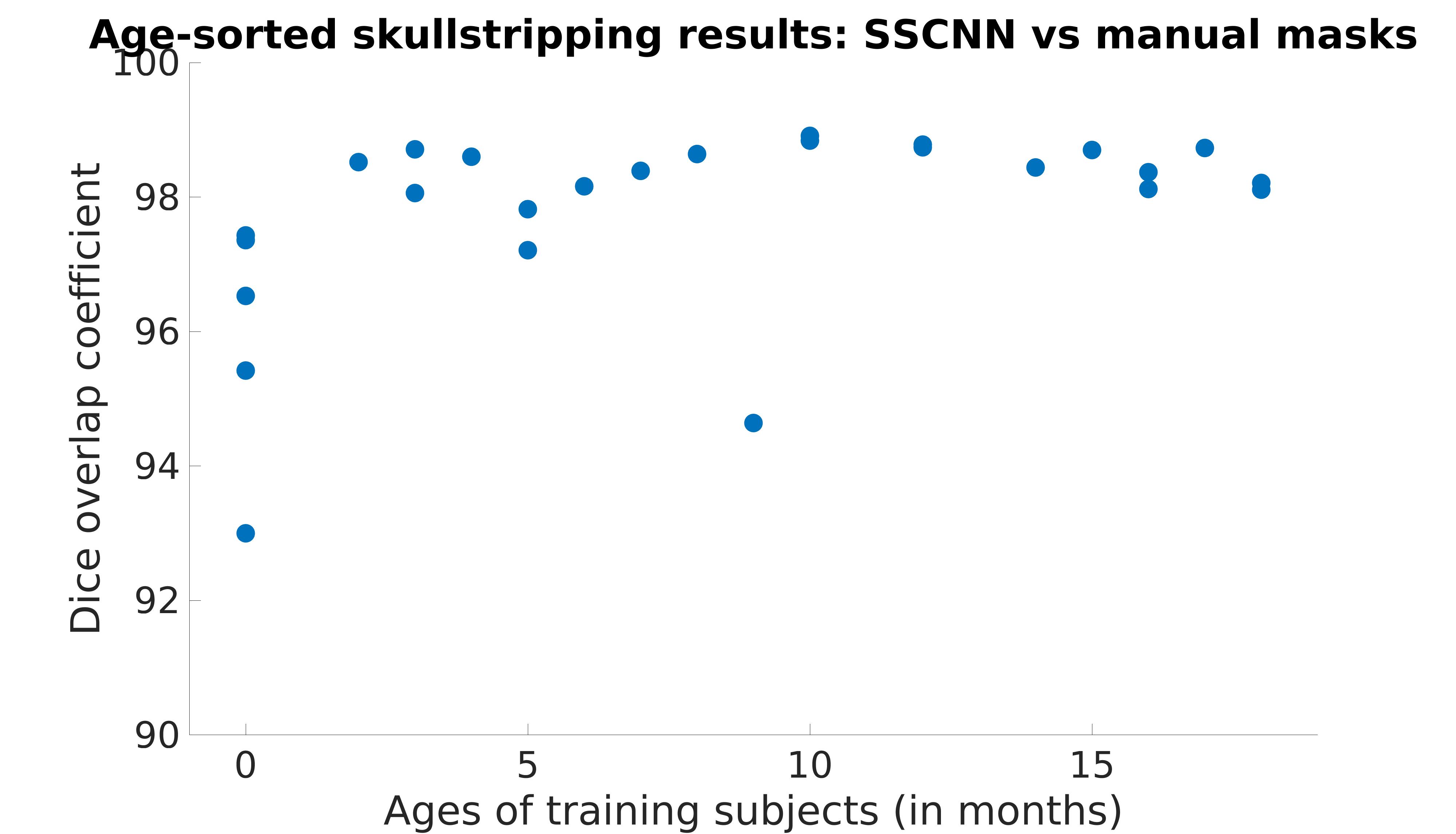}
    \caption{Dice overlap coefficients on the training data using our new skullstripping tool in a LOOCV framework}
    \label{fig:training_overlap_SSCNN}
\end{figure}

We chose the best network with parameters selected from the experiments described in Section~\ref{sec:parameters}. We ran SSCNN in a leave-one-out-cross-validation framework (testing data was excluded from the training data) and evaluated the outcomes with the given ground truth segmentation using Dice and Jaccard overlap metrics. The mean and standard deviation of these overlap measures were: for Dice $(97.79;1.43)$ and for Jaccard $(95.7;2.66)$. Figure \ref{fig:training_overlap_SSCNN} displays the individual Dice score results.

In order to put such a performance into perspective, we selected a set of five publicly available and widely used skullstripping algorithms and evaluated them on the same data set. These tools were: ROBEX, BET, BSE, 3dSkullStrip and Watershed. We also optimized (where appropriate) the key parameters of these tools on the training data set, in order to run comparison as fair as possible. In the below description, we indicate the parameters, their optimization range and the optimized parameters.

\underline{Skullstripping tools}: The RObust Brain Extraction \textbf{(ROBEX)}\cite{iglesias2011robex} tool had been primarily designed for adult input images. It deforms a brain surface to brain boundary, which is found by the brain versus non-brain classification. The deformed brain surface is then locally refined by a graphcut algorithm to obtain the final brain mask. ROBEX had performed well among 10+ skullstripping algorithms in multi-site brain images of the adults \cite{iglesias2011robex} and in children as young as newborns\cite{Serag:2016}. The advantage of ROBEX is that it does not require users to modify any parameters.
 The Brain Extraction Tool \textbf{(BET)} \cite{smith2002BET} is part of the FSL image analysis pipeline. It was primarily designed for images of adults. It evolves a deformable model to fit the brain surface. The performance of BET is often sensitive to parameter variations \cite{Lee:2003,Popescu:2012}. BET was used for skullstripping neonatal brain images \cite{shi2012label, Makropoulos:2014, Serag:2012}; but only upon careful and manual parameter tuning \cite{Popescu:2012}. We varied the fractional intensity threshold (f) and vertical gradient in fractional intensity threshold (g) parameters: f:[.2:.05:.8], g:[-.3:.05,.3]; (f,g)=(-.8,.05)
 Brain Surface Extractor \textbf{(BSE)} \cite{shattuck2001bes} had been primarily designed for images of adults. It smooths the input image and uses edge detectors to find brain boundary, and refines the results by morphological operations. BSE was used in \cite{shi2012label, Serag:2016} for neonatal images upon parameter tuning. We varied the diffusion (d) and edge detection constants (s): d:[10:5:60], s:[.42:.04:.82]; (d,s)=(-10,.58).
\textbf{3dSkullStrip} is part of the AFNI image analysis pipeline \cite{cox1996afni}. It replaces BET's deformable model with a spherical surface expansion paradigm, and modifies BET in other parts to avoid the eyes and to reduce leakage into the skull. We varied the brain vs non-brain intensity threshold (shrink\_fac) and speed of expansion (exp\_frac) parameters: shrink\_fac: [.4:.05:.8], exp\_frac: [.05:.025:.15]; (shrink\_fac, exp\_frac)= (0.8, .05).
Hybrid Watershed \textbf{(WATERSHED)} \cite{segonne2004hybrid} is part of the FreeSurfer neuroimaging analysis pipeline. It creates an initial brain surface using a watershed algorithm, and then evolves the brain surface to refine the result. We varied the preflooding height (h) parameter: h:[10:2:40]; h= 10.

\begin{figure}[!htbp]
    \centering
    \includegraphics[width=0.8\textwidth]{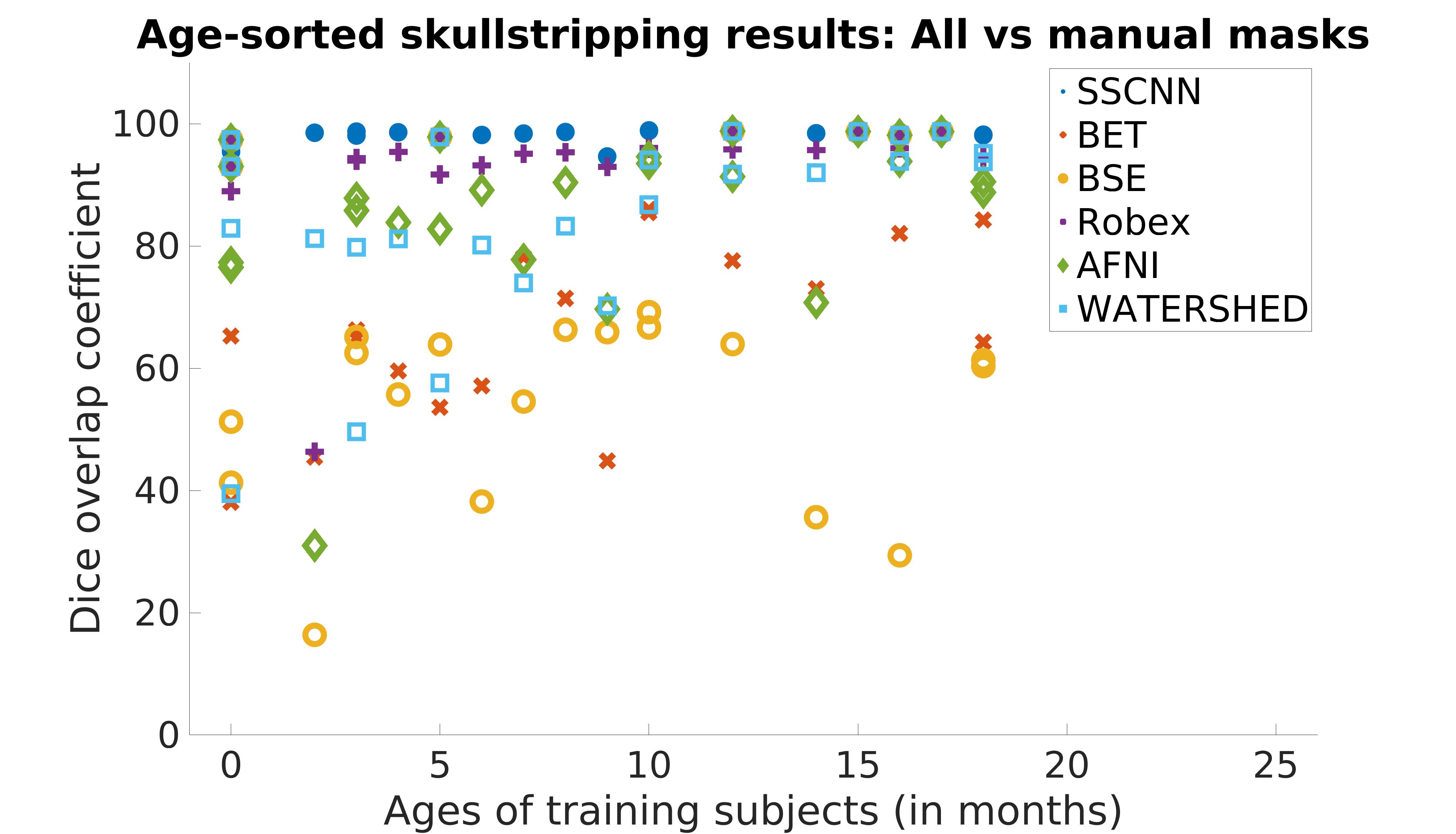}
    \caption{Dice overlap coefficients on the training data using all methods with their optimal flags}
    \label{fig:training_overlap_all}
\end{figure}

The mean and standard deviation of these overlap measures are included in Table \ref{tab:dice_and_jaccard_all} and Figure \ref{fig:training_overlap_all} displays the individual Dice score results.

\begin{table}[]
\begin{center}
\begin{tabular}{lcc}
 Method & Dice (Mean; std. dev.) & Jaccard (Mean; std. dev.) \\\hline
\textbf{SSCNN} & (97.8; 1.4) &  (95.7; 2.6) \\
\textbf{BET} & (76.0; 18.9) &  (64.9; 24.6) \\
\textbf{BSE} & (67.2; 24.1) & (55.6; 29.0) \\
\textbf{ROBEX} & (93.4; 9.8) & (88.7; 12.6) \\
\textbf{AFNI} & (86.7; 14.3) & (78.7; 18.2) \\
\textbf{Watershed} & (84.9; 15.7) & (76.4;20.7)\\\hline
\end{tabular}
\caption{Mean and standard deviation of Dice and Jaccard overlap metrics computed on the training data set with SSCNN and five other commonly used skullstriping techniques.}
\end{center}
\label{tab:dice_and_jaccard_all}
\end{table}

\subsubsection{Evaluation on an Independent Newborn Data Set}
\label{sec:uct_accuracy}
Anonymized brain MRI images of eighteen newborns from a cohort of 43 non-sedated infants born to 32 heavy drinkers and 11 controls recruited prospectively during pregnancy for a brain imaging study \cite{Jacobson:2017} was selected and manually traced to be used as a second data set to quantitatively evaluate the performance of our new skullstripping algorithm. For imaging details, see Table \ref{tab:table_test_data sets}. The segmenters were blind with respect to the newborn fetal alcohol spectrum disorder diagnosis and prenatal alcohol and drug exposure.
The mean and standard deviation of the Dice and Jaccard overlap scores was : $(97.51, 0.35)$ and  $(95.15, 0.67)$.

% as displayed on Figure \ref{fig:uct_overlap}.
%
%\begin{figure}[!htbp]
%    \centering
%    \includegraphics[width=0.45\textwidth]{figs/UCT_manual_vs_DL_dice_scores.jpg}
%    \includegraphics[width=0.45\textwidth]{figs/UCT_manual_vs_DL_jaccard_scores.jpg}
%    \caption{Dice and Jaccard overlap coefficients on the UCT test data using our new %skullstripping tool}
%    \label{fig:uct_overlap}
%\end{figure}
%
%
\begin{landscape}

\begin{table}[]
\caption {Test Data Set Summary}
\label{tab:table_test_data sets}

\resizebox{\linewidth}{!}{%
\begin{tabular}{ccccccl}

\cline{1-6}
\multicolumn{1}{|c|}{\textbf{Data Set}} & \multicolumn{1}{c|}{\textbf{N}} & \multicolumn{1}{c|}{\textbf{Age-at-scan}} & \multicolumn{1}{c|}{\textbf{Parameters}} & \multicolumn{1}{c|}{\textbf{Scanner}} & \multicolumn{1}{c|}{\textbf{MRI sequence}} &  \\ \cline{1-6}
\multicolumn{1}{|c|}{\textbf{dHCP}} & \multicolumn{1}{c|}{40} & \multicolumn{1}{c|}{Term age (37-44 weeks)} & \multicolumn{1}{c|}{\begin{tabular}[c]{@{}c@{}}T1wTR = 4795ms; TI = 1740ms; TE = 8.7ms;\\   SENSE factor 2.27 (axial) and 2.66 (sagittal);\\   0.8x0.8mm\textasciicircum{}2 and 1.6mm slices overlapped by 0.8mm\end{tabular}} & \multicolumn{1}{c|}{3T Philips Achieva} & \multicolumn{1}{c|}{inversion recovery T1w multi-slice fast spin-echo} &  \\ \cline{1-6}
\multicolumn{1}{|c|}{\textbf{BCH1}} & \multicolumn{1}{c|}{111+88} & \multicolumn{1}{c|}{10-52 and 77-156 days (avg=26.66,std=9.18 and avg=104.8, std=16.18)} & \multicolumn{1}{c|}{\begin{tabular}[c]{@{}c@{}}TR = 250 ms, TE=1.74 ms, TI = 1450ms, flip angle=7°, PAT=2, 1mm3\\   voxels, FOV=160mm\end{tabular}} & \multicolumn{1}{c|}{3T Siemens Trio} & \multicolumn{1}{c|}{\begin{tabular}[c]{@{}c@{}}T1-weighted mocoMEMPRAGE\\   {[}17,\\   18{]}\end{tabular}} &  \\ \cline{1-6}
\multicolumn{1}{|c|}{\textbf{BCH2}} & \multicolumn{1}{c|}{29+6} & \multicolumn{1}{c|}{\begin{tabular}[c]{@{}c@{}}1-26, 106-202 days\\   (avg=6.1, std=6.2 and avg=164.5, std=41.1)\end{tabular}} & \multicolumn{1}{c|}{\begin{tabular}[c]{@{}c@{}}TR = 250 ms, TE=1.74 ms, TI = 1450ms, flip angle=7°, PAT=2, 1mm3\\   voxels, FOV=160mm\end{tabular}} & \multicolumn{1}{c|}{3T Siemens Trio} & \multicolumn{1}{c|}{\begin{tabular}[c]{@{}c@{}}T1-weighted mocoMEMPRAGE\\   {[}17,\\   18{]}\end{tabular}} &  \\ \cline{1-6}
\multicolumn{1}{|c|}{\textbf{BCH3}} & \multicolumn{1}{c|}{105} & \multicolumn{1}{c|}{2-19 days (avg=9.32, std=3.59)} & \multicolumn{1}{c|}{\begin{tabular}[c]{@{}c@{}}TR = 2270 ms, TI = 1450\\   ms, flip angle = 7 degrees, 176 slices, 1 mm3 voxels, FOV =\\   220x220 mm2, GRAPPA=2\end{tabular}} & \multicolumn{1}{c|}{3T Siemens} & \multicolumn{1}{c|}{\begin{tabular}[c]{@{}c@{}}T1-weighted mocoMEMPRAGE\\   {[}17,\\   18{]}\end{tabular}} &  \\ \cline{1-6}
\multicolumn{1}{|c|}{\textbf{BAN}} & \multicolumn{1}{c|}{54} & \multicolumn{1}{c|}{\begin{tabular}[c]{@{}c@{}}62-97 days (avg=79.79,\\   std=9.26)\end{tabular}} & \multicolumn{1}{c|}{\begin{tabular}[c]{@{}c@{}}TR = 2520 ms, TE = 2.22\\   ms, 144 sagittal slices, 1 mm3 voxels, FOV = 192 mm\end{tabular}} & \multicolumn{1}{c|}{3T Siemens Verio} & \multicolumn{1}{c|}{T1-weighted MPRAGE} &  \\ \cline{1-6}
\multicolumn{1}{|c|}{\textbf{UCT}} & \multicolumn{1}{c|}{18} & \multicolumn{1}{c|}{\begin{tabular}[c]{@{}c@{}}7-47 days (avg=17.4,\\   std=11.5)\end{tabular}} & \multicolumn{1}{c|}{\begin{tabular}[c]{@{}c@{}}MEF 5°/20°: TR =\\   20ms, 8 echoes TE = 1.46 ms + n×1.68 ms where\\   n = 0,..,7, 144 sagittal slices, 1 mm3 voxels\end{tabular}} & \multicolumn{1}{c|}{3T Siemens Allegra} & \multicolumn{1}{c|}{\begin{tabular}[c]{@{}c@{}}multi-flip\\   angle, multi-echo FLASH (MEF)\end{tabular}} &  \\ \cline{1-6}
\multicolumn{1}{l}{} & \multicolumn{1}{l}{} & \multicolumn{1}{l}{} & \multicolumn{1}{l}{} & \multicolumn{1}{l}{} & \multicolumn{1}{l}{} &  \\
\multicolumn{1}{l}{} & \multicolumn{1}{l}{} & \multicolumn{1}{l}{} & \multicolumn{1}{l}{} & \multicolumn{1}{l}{} & \multicolumn{1}{l}{} &  \\
\multicolumn{1}{l}{} & \multicolumn{1}{l}{} & \multicolumn{1}{l}{} & \multicolumn{1}{l}{} & \multicolumn{1}{l}{} & \multicolumn{1}{l}{} &
\end{tabular}%
}
\end{table}

\end{landscape}

\subsubsection{Evaluation on Unlabeled Data Sets}
\label{sec:unlabeled_accuracy}
In addition to the previously described experiments, we also assembled a collection of anonymized and unlabeled test data set from five different initiatives in order to demonstrate the exceptional performance of our new tool on a variety of input images. Three of these originate from the Boston Children’s Hospital (BCH), one from Bangladesh and one from the recently released data set from the “The Developing Human Connectome Project” (dHCP) \cite{dhcp}. We compared the performance of SSCNN both qualitatively and quantitatively to the five already mentioned skull stripping solutions from Section \ref{sec:accuracy}.

Below is the description of all the data sets that were used for our experiments. The participating infants underwent structural MRI imaging and T1-weighted scans were acquired either on a 3T Siemens or a Phillips scanner during natural sleep. Human subject approval was obtained from all respective Institutional Review Boards and written informed parental consent was obtained for imaging. For details of the imaging protocol refer to Table \ref{tab:table_test_data sets}.

\begin{figure}[!htbp]
    \centering
    \includegraphics[width=0.85\textwidth]{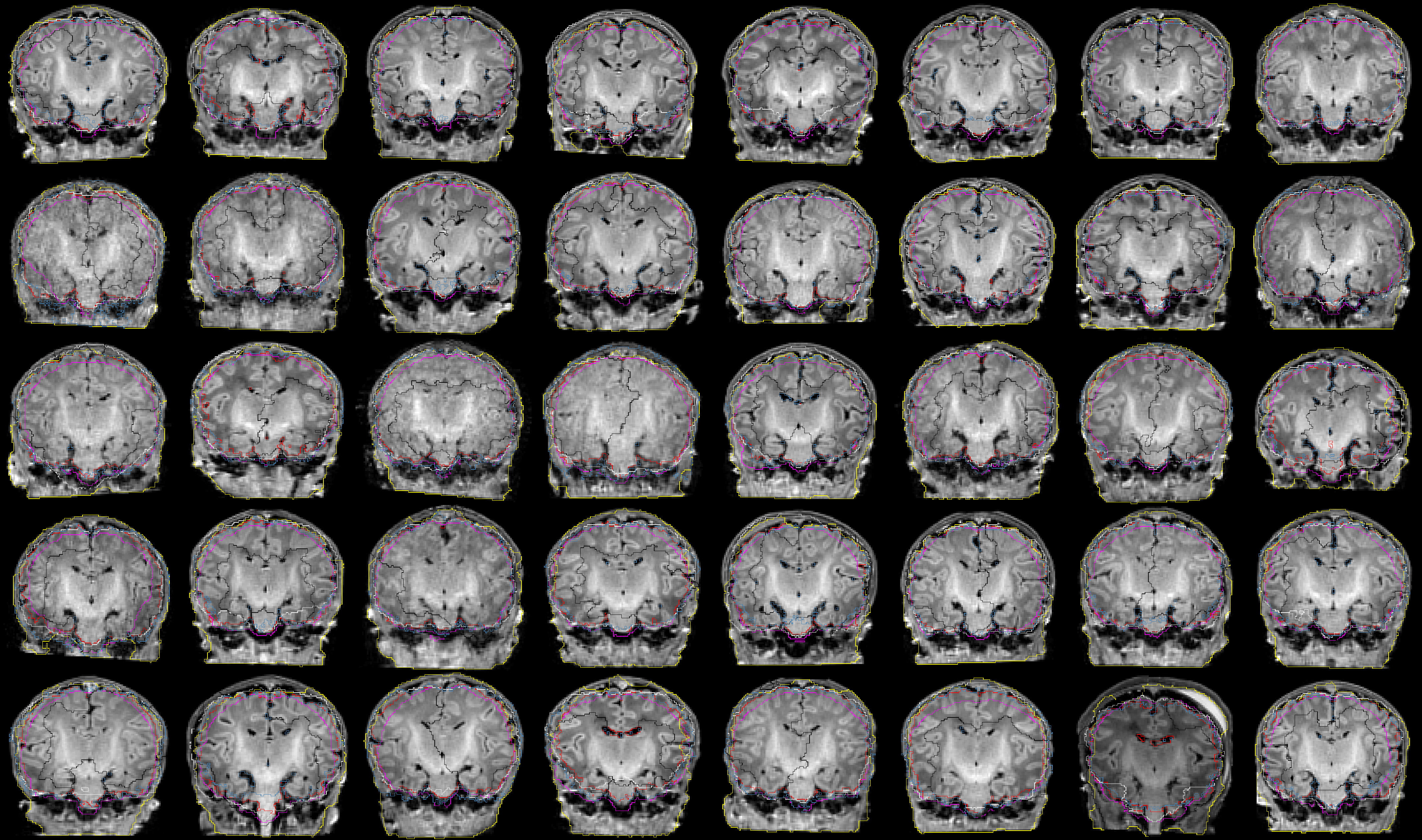}
    \caption{Skullstripping solutions on all 40 data sets from the dHCP project using six different tools. The images are aligned in an unbiased affine coordinate space for visualization purposes and the central coronal slice is selected from each of the MRIs. Skullstripping contours are indicated in color: BET (pink), BSE (yellow), robex (white), AFNI (blue), watershed (black) and SSCNN (red)}
    \label{fig:DHCP_all_skullstrippers_inCommon}
\end{figure}

\begin{figure}[!htbp]
    \centering
    \includegraphics[width=0.85\textwidth]{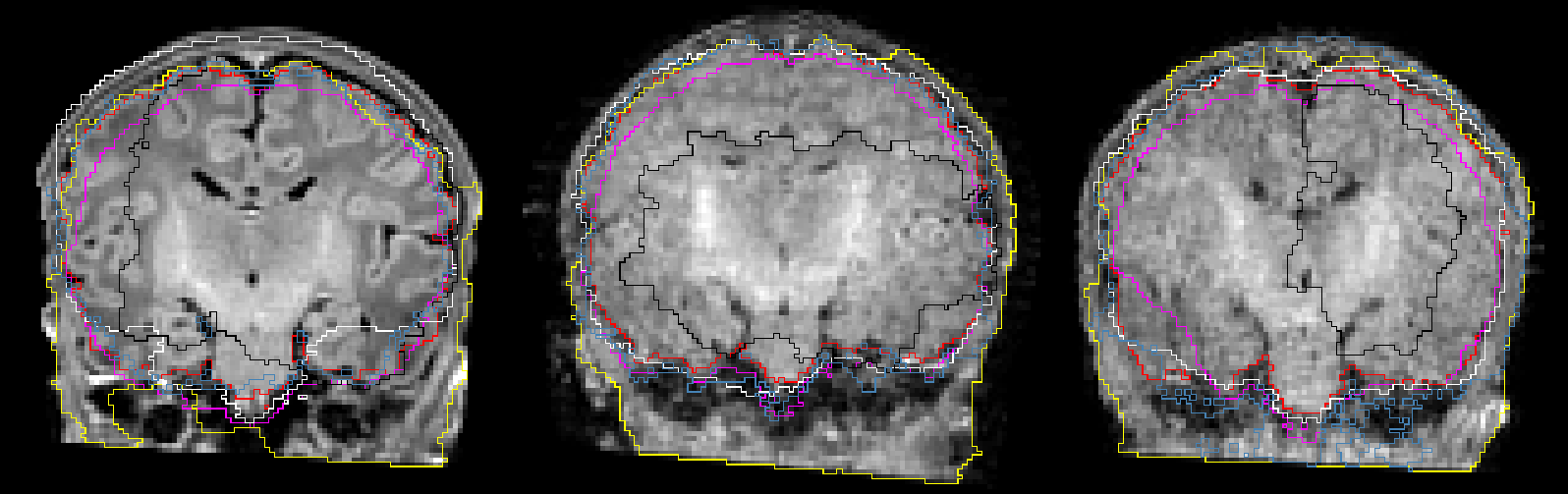}
    \caption{Skullstripping solutions on three representative data sets from the dHCP project using six different tools. The images are aligned in an unbiased affine coordinate space for visualization purposes and the central coronal slice is selected from each of the MRIs. Skullstripping contours are indicated in color: BET (pink), BSE (yellow), robex (white), AFNI (blue), watershed (black) and SSCNN (red)}
    \label{fig:DHCP_all_skullstrippers_inCommon_3examples}
\end{figure}

\begin{figure}[!htbp]
    \centering
    \includegraphics[width=0.85\textwidth]{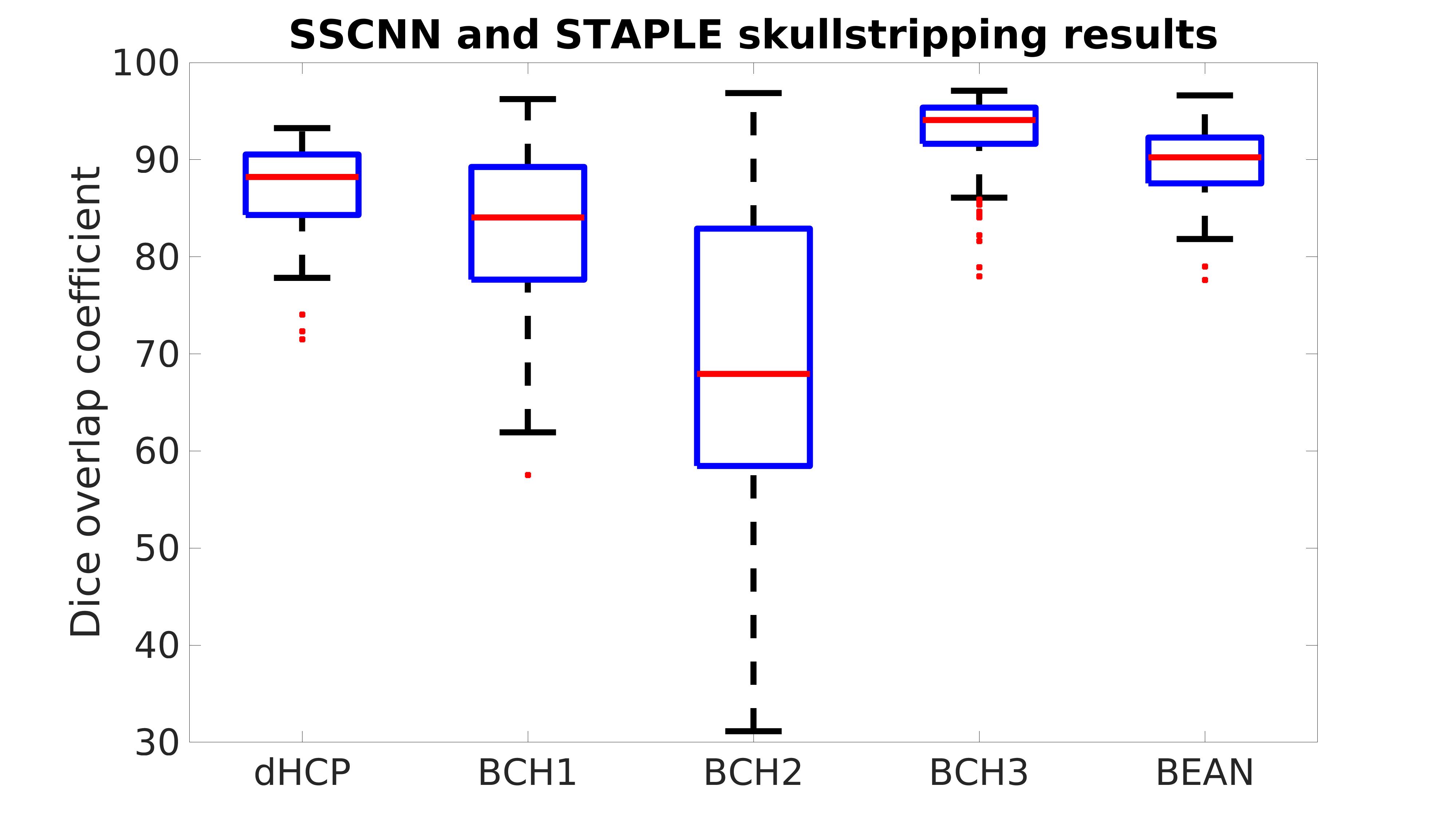}
    \caption{Dice overlap coefficients on unlabeled test data between solutions of our tool and the STAPLEd version of all tested automated algorithms}
    \label{fig:dl_vs_staple}
\end{figure}

\underline{Unlabeled Data Sets}:
\textbf{dHCP} $(N=40)$: Imaging data of forty newborn subjects was released by the developing human connectome project \cite{dhcp}. Even though the consortium processed the T2w images of these subjects in their initial release, the corresponding T1w images were also made available and we processed these in our study. The original data sets of 0.8 x 0.8 x 0.8 $mm\textsuperscript{3}$ were downsampled to 1mm isotropic for our processing.
\textbf{BCH1} $(N=199)$: Healthy, full-term neonates were recruited at the Brigham and Women’s Hospital (BWH) and Beth Israel Deaconess Medical Center (BIDMC) as part of an ongoing prospective data collection study. The protocol was reviewed and approved by the institutional review boards at Boston Children’s Hospital (BCH), BWH and BIDMC. The participating infants were all singletons with normal Apgar scores and have no clinical concerns regarding perinatal brain injury or congenital or metabolic abnormalities. All subjects were full-term infants scanned within their first month of life and a subset called back for a second scan at about 4 months of age.
\textbf{BCH2} $(N=35)$: Parents of neonates with congenital heart disease were approached for consent in the Cardiac ICU at BCH. This prospective study was approved by the institutional review board of BCH and was performed in compliance with the Health Insurance Portability and Accountability Act. Criteria for inclusion were: diagnosis of CHD confirmed by echocardiogram or cardiac MRI and ability to safely tolerate the brain MRI examination without sedation, prior and in some cases post to surgery. Neonates were excluded if there was evidence of a syndrome or genetic disease.
\textbf{BCH3} $(N=105)$: Native English-speaking children with and without a family history of DD were studied. All children were enrolled in a longitudinal dyslexia study which was approved by BCH institutional review.  The data set used in this study is from the first timepoint acquisitions.
\textbf{BEAN} $(N=54)$: Imaging data were collected in a set of infants from the Bangladesh Early Adversity Neuroimaging (BEAN) study investigating the effect of early biological and psychosocial adversity on children’s neurocognitive development among infants and children growing up in Dhaka, Bangladesh~\cite{Jensen:2019,Storrs:2017}.
For the set of unlabeled data sets, we used the optimized version of SSCNN and the five other skullstripping tools. Figures \ref{fig:DHCP_all_skullstrippers_inCommon} and \ref{fig:DHCP_all_skullstrippers_inCommon_3examples} display outcomes on the dHCP data, where SSCNN clearly outperforms the rest. Given that we did not have access to a ground truth solution, we STAPLEd \cite{Warfield:2004} all the outcomes together and then compared the SSCNN solution to it. The expectation was that the more SSCNN outperformed the other solutions the lower the Dice overlap score was between it and the STAPLEd labels, as well as the higher standard deviation. Both of these are well demonstrated in Figure \ref{fig:dl_vs_staple}.

%% Figure colorcoding: DL: red; BET: pink, BSE: yellow; Robex: white; AFNI: blue, watershed: black
%

\section{Discussion and Conclusion}
\label{sec:conclusion}
We have described SSCNN, a 2D multi-view CNN-based skullstripping method
for infant MRI. SSCNN trains three independent networks to extract
the brain mask from slices in three cardinal orientations--coronal, axial, and saggital. The outputs of the three networks are linearly combined to produce
a final brain mask. We have demonstrated in Section~\ref{sec:accuracy}
and Section~\ref{sec:uct_accuracy} that SSCNN is a highly accurate
skullstripping algorithm and is significantly better than existing methods.

We ran SSCNN on hundreds of subjects from diverse, multi-site, multi-scannner
MRI studies~(Section~\ref{sec:unlabeled_accuracy})and showed an improved
skullstripping performance compared to five other tools. SSCNN did not have
a single case of gross skullstripping failure on any of these datasets,
demonstrating its generalizability and robustness. Choosing a training
dataset encompassing the infant age range ensured that SSCNN was robust to the
contrast changes due to brain development.

SSCNN is computationally fast with a run time of less than 30 seconds with
a GPU~(NVidia P100, P40, Quadro P6000) and less than 2 minutes on a single
thread CPU. It is about 10 times faster than
ROBEX, which is the fastest method among those tested. This is an
important advantage as SSCNN can be potentially deployed on the scanner
to quickly identify the brain during a scanning session and perform slice
prescription for an optimal field of view based on the brain structure of
interest. This will also prove to be useful for motion correction between
scans--a step that is sometimes necessary when scanning infant subjects.

Presently, SSCNN is designed to work on $T_1$-weighted acquisitions. In the
next version we plan to update the training with cross-sequence augmentation
that will enable to it skullstrip acquisitions with $T_2$-weighted contrasts
as well~\cite{Jog:2019}. The linear combination of the three predictions in coronal, axial,
and sagittal orientations was learned from cross-validation experiments.
In the future, we will learn the weights of this combination by adding a custom layer that will collects the slice predictions in each orientation, reorients them, and combines them. This will ensure an end-to-end learning scheme for
combining all three orientations and jointly optimizing their weights.

We also need to evaluate SSCNN for infant subjects with pathologies
such as tumors, large ventricles that can significantly change the brain
shape and boundary characteristics of brain and skull. Our training dataset
does not have such subjects and we would need to augment the existing
training dataset or add newer training datasets to enhance this feature for
SSCNN.

In summary, we have described SSCNN, a fast, robust, infant MRI skullstripping
framework. The code will be made available as a part of the FreeSurfer development version repository~(\url{https://github.com/freesurfer/freesurfer}).
Further validation and testing will be necessary before incorporating it into
a release version.

%
%
%
%
%
%
% %% The Appendices part is started with the command \appendix;
% %% appendix sections are then done as normal sections
% %% \appendix
%
% %% \section{}
% %% \label{}
%
% %% References
% %%
% %% Following citation commands can be used in the body text:
% %% Usage of \cite is as follows:
% %%   \cite{key}          ==>>  [#]
% %%   \cite[chap. 2]{key} ==>>  [#, chap. 2]
% %%   \citet{key}         ==>>  Author [#]
%
% %% References with bibTeX database:
%
%
\section{Acknowledgements}
Support for this research was provided in part by the BRAIN Initiative Cell
Census Network grant U01MH117023, the National Institute for Biomedical Imaging
and Bioengineering (P41EB015896, 1R01EB023281, \\R01EB006758, R21EB018907,
R01EB019956), the National Institute on Aging (5R01AG008122, R01AG016495), the
National Institute of Mental Health, the National Institute for Neurological
Disorders and Stroke (R01NS0525851, R21NS072652, R01NS070963, R01NS083534,
5U01NS086625,\\ 5U24NS10059103), the Eunice Kennedy Shriver National Institute of Child Health \& Human Development (5R01HD065762), the National Institute on Alcohol Abuse and Alcoholism (R21 AA020037) and was made possible by the resources provided by Shared
Instrumentation Grants 1S10RR023401, 1S10RR019307, and 1S10RR023043. Additional
support was provided by the NIH Blueprint for Neuroscience Research
(5U01-MH093765), part of the multi-institutional Human Connectome Project. In
addition, BF has a financial interest in CorticoMetrics, a company whose medical
pursuits focus on brain imaging and measurement technologies. BF's interests
were reviewed and are managed by Massachusetts General Hospital and Partners
HealthCare in accordance with their conflict of interest policies.

\bibliographystyle{model1-num-names}
\bibliography{ms}

%% Authors are advised to submit their bibtex database files. They are
%% requested to list a bibtex style file in the manuscript if they do
%% not want to use model1-num-names.bst.

%% References without bibTeX database:

% \begin{thebibliography}{00}

%% \bibitem must have the following form:
%%   \bibitem{key}...
%%

% \bibitem{}

% \end{thebibliography}

\end{document}